\documentclass[twocolumn]{article}
\usepackage[margin=1.2cm,top=.5cm, bottom=2cm]{geometry}
\usepackage{cite}
\usepackage{amsmath,amssymb,amsfonts}
\usepackage{algorithmic}
\usepackage{graphicx}
\usepackage{textcomp}
\usepackage{bm}
\usepackage[table]{xcolor}
\def\BibTeX{{\rm B\kern-.05em{\sc i\kern-.025em b}\kern-.08em
    T\kern-.1667em\lower.7ex\hbox{E}\kern-.125emX}}

\usepackage{orcidlink}
\usepackage{booktabs}
\usepackage{cuted}

\newcommand{\FriW}[1]{\emph{FriWalk}}

\begin{document}

% Adapted from https://tex.stackexchange.com/questions/282418/onecolumn-titlepage-twocolumn-rest-of-document
\begin{strip}
\begin{center}
    {\huge Learning Priors of Human Motion With Vision Transformers} \vspace{4mm}
    \\
    Placido Falqueto$^1$\,\orcidlink{0000-0003-4233-5741},
    Alberto Sanfeliu$^2$\,
    \orcidlink{0000-0003-3868-9678},
    Luigi Palopoli$^1$\,
    \orcidlink{0000-0001-8813-8685}, 
    Daniele Fontanelli$^3$\,
    \orcidlink{0000-0002-5486-9989} \vspace{3mm} \\ 
    $^1$ Dept. of Information Engineering and Computer Science, University of Trento, Italy \\
%     Institut de Robòtica i Informàtica Industrial (CSIC-UPC)} \\
%     \textit{Universitat Politècnica de Catalunya}\\
%     Barcelona, Spain}\\
    $^2$Institut de Robòtica i Informàtica Industrial (CSIC-UPC), Universitat Politècnica de Catalunya, Barcelona, Spain\\
    $^3$ Dept. of Industrial Engineering, University of Trento, Italy
\end{center}
\end{strip} \vspace{2mm}

% \title{Learning Priors of Human Motion With Vision Transformers\\
% % \thanks{Identify applicable funding agency here. If none, delete this.}
% }
% 
% \author{
%     \IEEEauthorblockN{
%     Placido Falqueto\,\orcidlink{0000-0003-4233-5741}}
%     \IEEEauthorblockA{ \textit{
%     Dept. of Information Engineering and Computer Science} \\
%     \textit{University of Trento}\\
%     Trento, Italy}\\
%     \IEEEauthorblockN{
%     Alberto Sanfeliu\,\orcidlink{0000-0003-3868-9678}}
%     \IEEEauthorblockA{\textit{
%     Institut de Robòtica i Informàtica Industrial (CSIC-UPC)} \\
%     \textit{Universitat Politècnica de Catalunya}\\
%     Barcelona, Spain}\\
%     \IEEEauthorblockN{
%     Luigi Palopoli\,\orcidlink{0000-0001-8813-8685}}
%     \IEEEauthorblockA{\textit{
%     Dept. of Information Engineering and Computer Science} \\
%     \textit{University of Trento}\\
%     Trento, Italy} \\
%     \IEEEauthorblockN{
%     Daniele Fontanelli\,\orcidlink{0000-0002-5486-9989}}
%     \IEEEauthorblockA{\textit{
%     Department of Industrial Engineering} \\
%     \textit{University of Trento}\\
%     Trento, Italy}
% }
% 
% \date{}
% \maketitle

\begin{abstract}
    %%%%%%%%%%%%%%%%%%%%%%%
%A clear understanding of where humans are located in a scene and how they
%are moving in the near future is a key enable for robot navigation within human-populated environments.
%We propose a neural architecture based on visual transformers to provide this information. This solution can arguably capture spatio-temporal correlations more effectively than standard neural networks. In the paper, we describe our solution and show numeric evidence of its advantages.
A clear understanding of where humans move in a scenario, their usual paths and speeds, and where they stop, is very important for different applications, such as mobility studies in urban areas or robot navigation tasks within human-populated environments.
We propose in this article, a neural architecture based on Vision Transformers (ViTs) to provide this information. This solution can arguably capture spatial correlations more effectively than Convolutional Neural Networks (CNNs). In the paper, we describe the methodology and proposed neural architecture and show the experiments' results with a standard dataset. We show that the proposed ViT architecture improves the metrics compared to a method based on a CNN.
%%%%%%%%%%%%%%%%%%%%%%%
\end{abstract}


%%%%%%%%%%%%%%%%%%%%%%%
%%%%%% SECTIONS %%%%%%%
%%%%%%%%%%%%%%%%%%%%%%%

\section{Introduction}
% \revv{Provide an introduction to the integration of mobile robots into human environments, emphasizing the challenges and the need for predicting areas of heightened pedestrian activity. Mention the motivation for your study and the objectives you aim to achieve.}

An essential requirement for a mobile robot to be able to move within
a human-populated environment~\cite{Singamaneni2024} is its ability to
evaluate the human occupancy of the different areas of the environment
and to foresee their most likely direction of motion in the near
future.  This information is reconstructed by humans by a quick sight
of the scene and is instinctively used to identify the most convenient
and efficient path to follow.  Robots require a collection of
sophisticated algorithms to accomplish the same results.
%In this paper, we will concentrate on the problem of understanding and predicting human motion.
In this paper, we will concentrate on the problem of understanding
where people move in a scenario, which are their common trajectories
and speeds and where they stop. This information can be mainly used to
know the priors of human motion for different applications of robot
navigation tasks, whilst its application is envisioned in many
different fields of robotics. For instance, motion priors are of
paramount importance in production plants when robots, most probably
cobots, deal with cooperative and coordinated tasks with humans in
modern robotics cells in order to improve simultaneously efficiency
and safety.

%Incorporating mobile robots into human environments presents multifaceted challenges, %particularly in predicting areas of heightened pedestrian activity—a crucial aspect for %efficiently managing high-traffic zones. Unlike humans who instinctively discern typical walking %paths based on environmental cues, mobile robots rely on sophisticated algorithms to navigate %complex spaces.

%Following the research line started by Rudenko et al.~\cite{rudenko}, our method focuses on capturing walking patterns %through the use of static semantic maps, which marks a significant departure from the traditional reliance on robust people %trackers. 
Motivated by previous papers~\cite{rudenko} on the importance of understanding human motion in shared spaces, our approach takes on the challenges of predicting occupancy priors for walking individuals in unfamiliar locations by relying solely on the semantic information of the area.
Semantic maps allow us to break down the observed area into small parcels. The resulting network has a low complexity and
is suitable for producing real--time predictions within a small time horizon. The price to pay is the loss of the "big picture", i.e., on how the motion between the different areas is related.
Our proposed solution brings about an important advance in the state of the art proposing the adoption of novel architecture based on Vision Transformers (ViTs) to predict the occupancy distributions of walking individuals. 
The choice of ViTs is dictated by their well-known ability to extract effectively contextual information. This feature
is exploited to understand the spatial relation between the different parcels, thereby enabling the network to 
reconstruct global information and learn how humans use the different areas (affordance). 
The resulting algorithm keeps the real--time computation cost within acceptable bounds, but it significantly
improves the performance of the network even in the face of quick changes in the environment.  

The simulation results unequivocally demonstrate that our ViT-based models outperform the baseline in terms of accuracy, reinforcing our belief that this solution can be a natural choice for real-world applications, in which mobile robots navigate across complex and dynamic environments.

The paper is organised as follows. In Section~\ref{sec:related}, we offer a thorough review of the state-of-the-art on existing methodologies for inferring occupancy prior distributions in semantically rich urban environments. 
In Section~\ref{sec:method}, we describe the key components of our proposed architecture, along with the proposed evaluation metrics and a description of the data set used in the training phase. In Section~\ref{sec:ablation}, we propose an ablation study to point out the impact of the different components in the architecture. In Section~\ref{sec:results}, we illustrate our simulation results on known data set to show the improvement brought by our solution over the baselines.
Finally, in Section~\ref{sec:conclusions}, we offer our conclusions and announce future work directions.
\section{Related Work}
\label{sec:related}

In this section, we explore the existing body of research on human
motion prediction, focusing particularly on the crucial role of map
priors inference. We provide a succinct overview of various Neural
Network architectures used in vision.  Additionally, we scrutinize the
methodologies and limitations of previous approaches, with a detailed
examination of Rudenko et al.'s {\tt semapp}~\cite{rudenko}. Notably,
we highlight the scarcity of literature addressing the direct
prediction of priors from maps, a gap we aim to fill with our work.

\subsection{Human Motion Prediction and Prior Occupancy Inference}

Anticipating human motion intentions represents a longstanding
challenge, demanding a nuanced comprehension of social dynamics~\cite{mavrogiannis2023core}.  As
described in the survey~\cite{rudenko2020human}, the modelling of human
motion trajectories can be categorized through the representation of the
underlying causes. Physics-based methods rely on explicit dynamical
models derived from Newton's laws, either with a single model or a set
of adaptive multi-models~\cite{FarinaFGGP17,7535484,7995734}. Pattern-based techniques
learn motion patterns from observed data, either sequentially over
time or non-sequentially considering the entire trajectory
distribution~\cite{vemula2017modeling}. Planning-based methods explicitly consider the agent's
long-term goals~\cite{AntonucciPBPF21access}, classifying into forward
planning, assuming explicit optimality criteria, and inverse planning,
estimating reward functions from observed trajectories.
In~\cite{rudenko2020human}, a significant increase in related works in
this area is described, particularly in pattern-based methods.

\subsubsection*{Importance of Prior Occupancy Inference}
Predicting prior occupancy distribution, rather than individual
trajectories, proves valuable in extrapolating contextual information
and enriching our understanding of a location. While the problems may
appear similar, they represent distinct perspectives. The former
focuses on dynamic predictions of individual or group actions within
an environment, while the latter involves analyzing the environment
itself, offering insights into typical human behaviours within that
context. This differentiation enhances our ability to anticipate
future events and make informed decisions based solely on
environmental information~\cite{kaleci2020semantic}. Despite the evident importance of this
approach, there is a distinct gap in existing literature dedicated to
the direct prediction of priors from maps.

\subsection{Neural Network Architectures}
We'd like to highlight that our literature review will prominently
showcase segmentation models, underscoring the extensive research in
this domain. Unlike classification tasks, segmentation
involves pixel-wise classification, where the goal is to assign a
class label to each pixel in an image, effectively creating segments
based on pixel content.  As you explore further sections, you will
observe our focus on a similar pixel-wise classification task, where
the objective is to predict the likelihood of human presence in
individual pixels. This focus aligns with cutting-edge approaches in
segmentation, known for generating output tensors with
the same dimension as the input. Consequently, our architecture is
designed to meet the unique requirements of these tasks.

\subsubsection*{Convolutional Neural Network (CNN)}
CNNs have been instrumental in the advancements in computer vision,
notably excelling in challenges such as the ImageNet classification
challenge~\cite{cnn}. Their success in image classification and
semantic segmentation has positioned them as a popular choice for
various vision-related tasks.  In spite of considerable efforts and
substantial advancements in recent years, image segmentation remains a
formidable challenge~\cite{minaee2021image}. This difficulty arises from the intricate
intra-class variations, contextual disparities, and ambiguities
introduced by occlusions and low image resolution. All these
limitations can be extended to the inference of map priors, where
context is key for a good prediction. In~\cite{doellinger}, Doellinger
et al. use the architecture introduced by Jegou et
al.~\cite{densecnn}, which combines dense blocks~\cite{denseblock} and
fully CNNs, to predict average occupancy maps of walking humans even
in environments where no human trajectory data is available, using the
static grid maps as input.  In~\cite{rudenko}, Rudenko et al. extend
this architecture using as input semantic maps, instead of the plain
grid map.

However, the landscape is undergoing a transformation, with a
discernible shift towards transformer-based models. While CNNs have
long dominated, the current trend indicates a growing adoption of
transformer architectures in various vision-related tasks. This shift
is clearly seen in the ImageNet classification challenge, where more
and more people are opting for Vision Transformers (ViTs)~\cite{srivastava2023omnivec, wortsman2022model}.

\subsubsection*{Vision Transformers}
Vision Transformers (ViTs)~\cite{vit} presents a cutting-edge approach
to image processing by incorporating self-attention mechanisms to
capture contextual information. While in image classification
often an encoder structure to downsample features into a
latent space and generate label predictions is enough, in image segmentation, we need to 
employ an encoder-decoder structure. In segmentation and
reconstruction tasks, this structure upsamples the latent space to
produce images with per-pixel class scores. To address the biases
towards local interactions observed in convolutional architectures
during segmentation tasks, Strudel et al.~\cite{strudel2021segmenter}
propose a novel perspective. They formulate semantic segmentation as a
sequence-to-sequence problem and adopt a transformer architecture to
leverage contextual information throughout the entire
model~\cite{vaswani2023attention}. The authors claim to surpass all
previous state-of-the-art convolutional approaches by a substantial
margin of $5.3\%$. This notable improvement is attributed, in part, to
the enhanced global context captured by their method at every layer of
the model.

The potential of ViTs in tasks related to human motion prediction remains an area of exploration. In this paper, we delve into the capabilities of Vision Transformers (ViTs) to map priors inference, investigating their applicability and performance in this domain.

\subsubsection*{Masked Autoencoder}
The success of masked language modelling, exemplified by
BERT~\cite{devlin2019bert} and GPT~\cite{brown2020language} in NLP
pre-training, lies in holding out portions of input sequences and
training models to predict the missing content. This method, proven to
scale excellently, has demonstrated effective generalization to
various downstream tasks. Inspired by these achievements, Masked
Autoencoders (MAEs)~\cite{mae} were developed to introduce a novel
approach in computer vision, specifically addressing challenges
related to latent representation learning.  In contrast to traditional
supervised learning in computer vision, which heavily depends on
labeled datasets, Masked Autoencoders (MAEs) adopt a self-supervised
approach for the classification task. Unlike conventional semantic
segmentation based on ViTs, where images are decomposed into visual analogs of
words, MAEs deviate by randomly removing patches during training. In
essence, MAEs focus on reconstructing pixels, which are not inherently
semantic entities. However, intriguingly, the MAE model demonstrates
the ability to infer complex and holistic reconstructions, suggesting
a learned understanding of various visual concepts and semantics. This
behavior hints at the presence of a rich hidden representation within
the MAE, leading to the hypothesis that the model captures diverse
visual concepts through its self-supervised learning framework.

In this paper, we extend its exploration beyond pixel
reconstruction. In particular, we delve into the performance of MAEs
in the realm of map priors inference, investigating their ability to
understand and interpret underlying visual concepts and semantics, and
comparing its performance to the ViT. By scrutinizing the model's
proficiency in this distinct task, we aim to unravel the extent to
which MAEs can harness their learned representations for more advanced
cognitive processes. This multifaceted analysis not only broadens our
understanding of MAEs in computer vision but also provides valuable
insights that can guide and inspire future research endeavors in the
field.

To pay homage to the pioneering work in~\cite{rudenko}, we
affectionately name our Vision Transformer-based approach "Semantic
Map-Aware Pedestrian Prediction 2" ({\tt semapp2}), described in the
next section

\section{Methodology}
\label{sec:method}

We start the description of the proposed solution by comparing the
different metrics to compute the distance between probability
distributions.

\subsection{Metrics}
\subsubsection{Kullback-Leibler divergence}
The Kullback-Leibler (KL) divergence~\cite{kl_divergence_paper} is a measure of how one probability distribution diverges from a second, expected probability distribution as
\begin{equation*}
  \text{KL}(P_{GT} || Q_{pred}) = \sum_i P_{GT}(i)
  \log\left(\frac{P_{GT}(i)}{Q_{pred}(i)}\right) ,
\end{equation*}
i.e., the divergence of the probability distribution of the prediction
$Q_{pred}$ from the probability distribution of the target $P_{GT}$,
over a discrete set of events indexed by $i$.

% The Kullback-Leibler (KL) divergence is often used in neural networks, particularly in probabilistic models and variational autoencoders (VAEs). In VAEs, KL divergence is commonly employed to measure the difference between the inferred probability distribution of latent variables and a predefined prior distribution. It helps regularize the latent space and encourages the learned distribution to be close to the desired prior.

The KL divergence is not symmetric, meaning that
$\text{KL}(P \, || \, Q)$ is not necessarily equal to
$\text{KL}(Q \, || \, P)$. In the context of neural networks training,
it is typically applied in the direction of the predicted distribution
($P$) compared to the target distribution ($Q$).  The reason for this
choice is often related to the nature of the optimization problem. In
tasks like probabilistic modeling or generative modeling, you want the
predicted distribution to approach or match the target
distribution. Minimizing the KL divergence in the direction of the
predicted distribution helps achieve this goal.

However, in this specific application, it might be meaningful to also
calculate the reverse KL divergence, i.e., $\text{KL}(Q \, || \, P)$,
contrary to the traditional machine learning approaches. This
unconventional choice can be justified by examining the KL divergence
formula: when $P_{GT}(i)$ is near zero, the KL divergence tends to be
low, potentially masking issues in predictions, leading to
misinterpretation of the model's performance. Instead, by calculating
the reverse KL divergence $\text{KL}(Q \, || \, P)$, the contribution
to the divergence is weighted based on the prediction, ensuring that
deviations in regions where the prediction is far from zero but the
target is zero are appropriately penalized.

\subsubsection{Earth Mover's Distance (EMD)}

The Earth Mover's Distance (EMD)~\cite{rubner2000earth}, also known as
Wasserstein distance or optimal transport distance, is a metric used
to quantify the dissimilarity between two probability
distributions. It provides a measure of the minimum amount of work
required to transform one distribution into another. More in-depth,
given two probability distributions \(P\) and \(Q\) representing the
histograms of pixel intensities in the occupancy distributions, and a
ground distance function \(d(x, y)\) representing the cost of
transporting mass from intensity \(x\) to intensity \(y\), the EMD is
defined as
\begin{equation*}
  \text{EMD}(P, Q) = \min_{\gamma \in \Gamma(P, Q)} \sum_{(x, y) \in \text{supp}(\gamma)} \gamma(x, y) \cdot d(x, y) .
\end{equation*}
Here, $\Gamma(P, Q)$ represents the set of all possible joint
distributions (couplings) of $P$ and $Q$ whose marginals are $P$ and
$Q$ respectively. The minimization is over these couplings, while
$\text{supp}(\gamma)$ denotes the support of the coupling, i.e., the
set of pairs $(x, y)$ with non-zero probability.

Unlike the KL divergence, the Earth Mover's Distance is a metric that
adheres to the triangle inequality and is symmetric. Its symmetry
makes it particularly suitable for scenarios where a balanced
evaluation of differences in both directions is desired. In the paper
experiments, we will employ the forward KL divergence (KL-div), the
reverse KL divergence (rKL-div) and the Earth Mover's Distance (EMD)
to thoroughly assess the performance of our model in capturing the
nuances of probability distributions.

\subsection{Datasets}
\label{sec:datasets}
% Describe the Stanford Drone Dataset (SDD), detailing its environments, semantic classes, and your enhancements. Provide insights into the preprocessing steps and visual representations.
% Provides details on data collection, preprocessing, and augmentation.

Our study builds upon the Stanford Drone Dataset
(SDD)~\cite{Robicquet2016}. This extensive dataset captures images and
videos featuring diverse agents like pedestrians, bicyclists,
skateboarders, cars, buses, and golf carts navigating a real-world
outdoor environment. It provides a comprehensive representation of
human motion in shared spaces.

% Specify which maps from stanford drone dataset where used for training and the fact that we use cross validation because we have little number of maps available.
We utilized a subset of 20 maps from the Stanford Drone Dataset for training, including "bookstore", "coupa", "death circle", "gates", "hyang", "little" and "nexus". Due to the limited availability of maps, we employed a cross-validation strategy, leaving one map out at a time for testing while training and validating on the remaining maps. This approach allowed us to maximize the use of the available data and ensure a robust evaluation of our model across various scenarios.

During preprocessing, the Stanford Drone Dataset (SDD) scenes were not only scaled but also manually segmented into refined semantic classes. All scenes were uniformly scaled to a resolution of 0.4 meters per pixel. Given that our network operates on map crops of fixed size ($64x64$ pixels), we adopted a strategy of decomposing the larger input images from the SDD into 500 random crops of appropriate size (like in~\cite{rudenko}). Each crop in the training data is augmented 5 times by rotating and mirroring. The final distribution \(p(s)\) for state \(s\) was reconstructed by averaging the predicted occupancy values of \(s\) across all crops containing that state. This approach, as stated in~\cite{rudenko}, enhances the robustness of our model's predictions by addressing potential artefacts associated with neighbouring crops.

To enhance prediction accuracy, we extend the semantic classes beyond those considered by Rudenko et al. \cite{rudenko}. In the paper, the authors use 9 semantic classes: pedestrian area, vehicle road, bicycle road, grass, tree foliage, building, entrance, obstacle and parking. We choose to add 4 more classes: sitting area, stairs, shaded area and intersection zone, reaching a total of 13 semantic classes.
We find that using semantic classes that heavily influence human motion greatly affects the accuracy of the predictions.
In Table~\ref{tab:semantic_influence_9} we compare the use of the 9 classes (pedestrian area, vehicle road, bicycle road, grass, tree foliage, building, entrance, obstacle and parking) and in Table~\ref{tab:semantic_influence_13} the complete model with all 13 classes (adding stairs, shaded area, intersection zone and sitting areas).
%-%
\begin{table}[t]
\footnotesize
  \centering
  \caption{Quantitative Evaluation of 9 Semantic Classes}
  \label{tab:semantic_influence_9}
  \begin{tabular}{lccc}
    \toprule
     & KL-div & rKL-div & EMD \\
    \midrule
    {\tt semapp} & $0.66\pm0.15$ & $2.50\pm1.51$ & $40.18\pm26.55$ \\
    {\tt semapp2} & $0.49\pm0.15$ & $2.15\pm1.20$ & $34.24\pm26.47$ \\
    \bottomrule
  \end{tabular}
\end{table}
%-%
%-%
\begin{table}[t]
\footnotesize
  \centering
  \caption{Quantitative Evaluation of 13 Semantic Classes}
  \label{tab:semantic_influence_13}
  \begin{tabular}{lccc}
    \toprule
     & KL-div & rKL-div & EMD \\
    \midrule
    {\tt semapp} & $0.58\pm0.14$ & $2.43\pm1.24$ & $41.16\pm26.98$ \\
    {\tt semapp2} & $0.46\pm0.16$ & $2.19\pm1.50$ & $27.65\pm19.89$ \\
    \bottomrule
  \end{tabular}
\end{table}
%-%
Notably, {\tt semapp2} already exhibits notable advancements in prediction
accuracy compared to {\tt semapp} when restricted to the original 9
semantic classes, as evidenced by the metrics in
Table~\ref{tab:semantic_influence_9}. In
Table~\ref{tab:semantic_influence_13}, {\tt semapp2} consistently
performs better than {\tt semapp} across all evaluation metrics. The
addition of the 4 new semantic classes refines the semantic
understanding and contributes to improved accuracy in predicting
occupancy distribution priors. This observation aligns with our goal
of enhancing the model's capability to capture nuances in
human-centric environments.

\subsection{Proposed Vision Transformer architecture}
% \revv{Describes the deep learning models, architectures, and algorithms used.
% Present the architecture details and key parameters of ViT. Describe the hyperparameter optimization strategies, including the leave-one-out approach, and the ablation study comparing different loss functions and patch sizes.}

\begin{figure}[t]
    \centering
    \includegraphics[width=\linewidth]{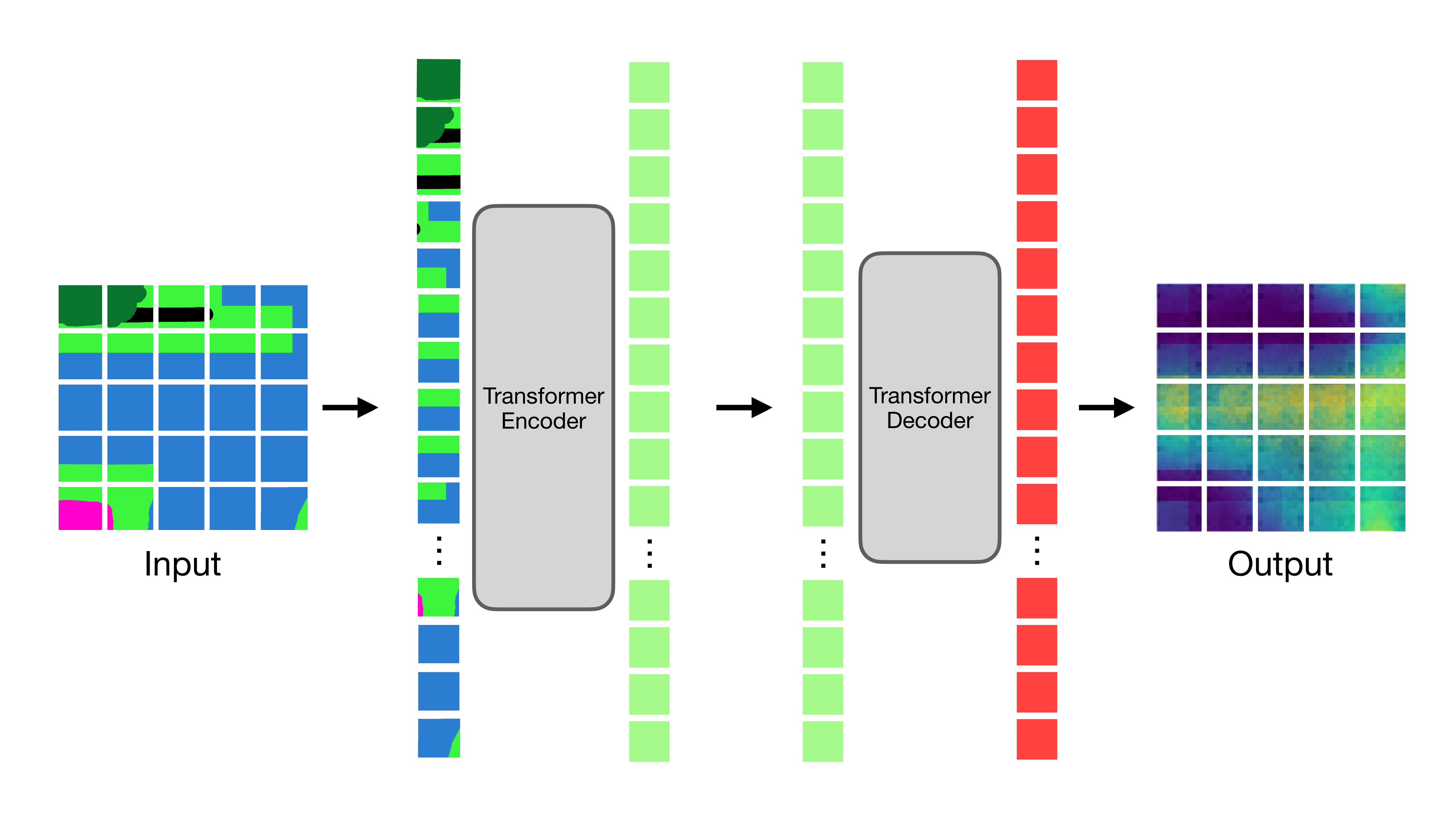}
    \caption{The {\tt semapp2} architecture.}
    \label{fig:semapp2arc}
\end{figure}

The proposed {\tt semapp2} consists of a ViT autoencoder designed to
generate a prior prediction image from the input semantic map with
multiple channels, each corresponding to different semantic classes
(see Figure~\ref{fig:semapp2arc}).
%-%
%-%
We use a simple autoencoder architecture, where an encoder maps the
observed signal (semantic map) to a latent representation, and a
decoder predicts the prior from the latent representation.

\begin{figure}[t]
    \centering
    \includegraphics[width=\linewidth]{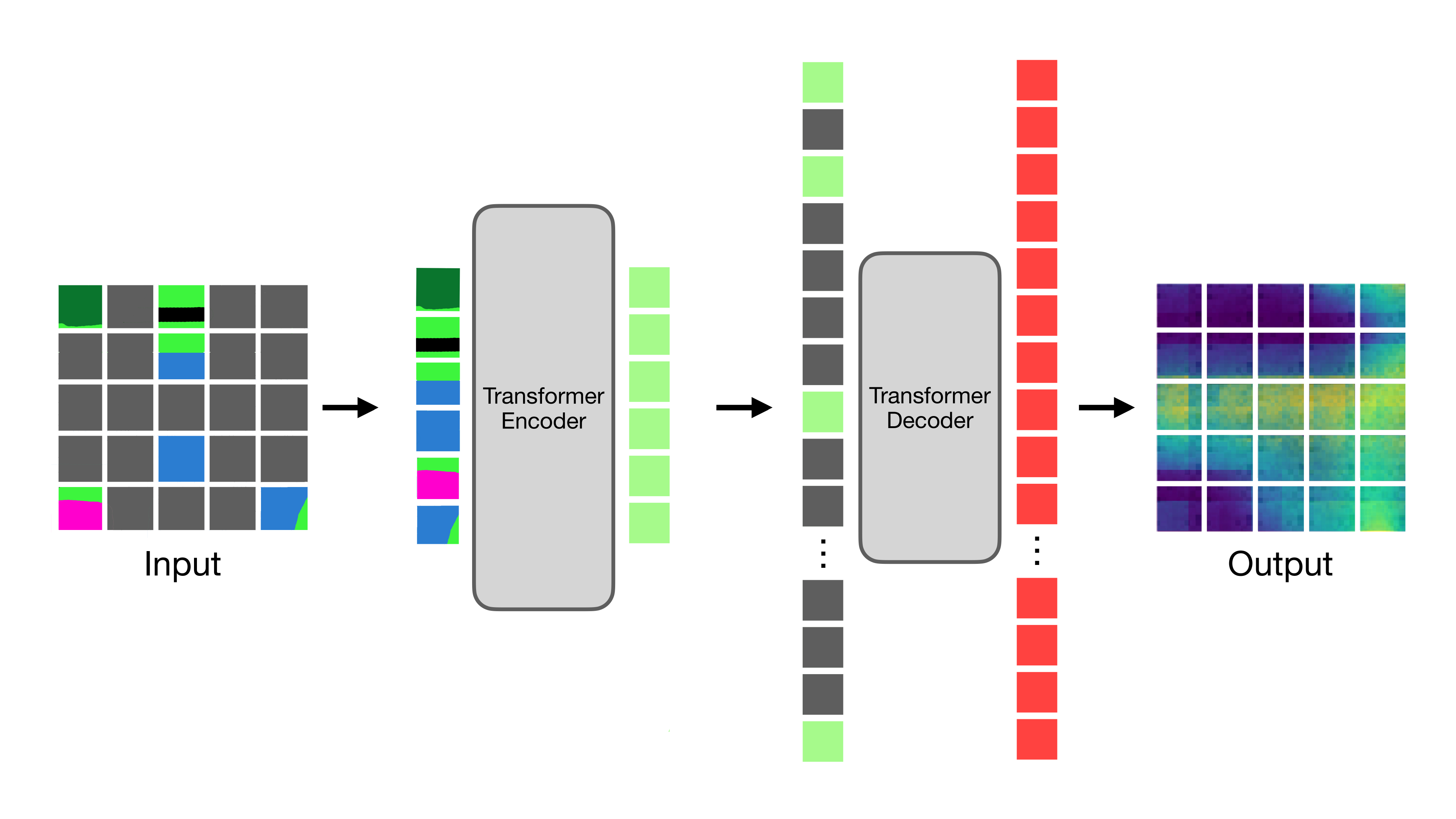}
    \caption{{\tt semapp2} variation using a MAE autoencoder.}
    \label{fig:semappMAE}
\end{figure}

In Figure~\ref{fig:semappMAE} we provide a visual representation of the {\tt MAE-semapp2}, an architecture variation of the {\tt semapp2} that uses a MAE autoencoder, with $75\%$ masking ratio.
Note that the two architectures are the same: if we set a masking ratio of $0\%$ on the {\tt MAE-semapp2}, we obtain the same behaviour of a ViT-based {\tt semapp2}.
For simplicity, we will refer to a {\tt semapp2} with a masking ratio of $75\%$ as {\tt MAE-semapp2} from now on.

\subsubsection*{Encoder}
The encoder employs the ViT architecture, customized for semantic map processing. The input semantic map, with multiple channels representing various semantic classes, undergoes a linear projection with added positional embeddings. Subsequently, the resulting set of tokens is processed through a series of Transformer blocks. In the MAEs variation, the encoder is identical to the ViT encoder, but it handles only the subset of unmasked patches of the semantic map.

\subsubsection*{Decoder}
The decoder takes the full set of tokens consisting of encoded visible patches, and mask tokens. Each semantic class in the map is represented by a learned vector, and positional embeddings are added to all tokens in the set for reconstruction purposes. The decoder consists of a series of Transformer blocks designed to reconstruct the prior prediction image. Notably, the decoder architecture is independent of the encoder's design, providing flexibility.

% Explains the experimental setup and evaluation metrics.}

\subsection{Training and Evaluation}
% Detail your primary objective of predicting prior occupancy distribution based on semantic information. Introduce the Vision Transformer (ViT) and semapp as comparison methods, and briefly touch on the Masked Autoencoder (MAE). Explain the quantitative evaluation metrics, such as KL-divergence and MSE.
Our primary objective is predicting prior occupancy distribution based
on semantic information, encompassing stop distribution and velocities
heat map prediction. This task expands previous works, such as Rudenko
et al.'s \cite{rudenko}, which focus on occupancy distribution
prediction only. We compare two main models in our study: our novel
framework {\tt semapp2}, which is based on ViTs, and Rudenko et al.'s
{\tt semapp}, based on CNNs. Additionally, we provide a concise
comparison with a variation of our {\tt semapp2}, based on the Masked
Autoencoder.

% Our quantitative evaluation relies on metrics such as the Kullback-Leibler divergence (KL-divergence) and Mean Squared Error (MSE) \pla{(Since the values are probabilities and being very small the overall MSE is very small)}, serving as measures of dissimilarity between the predicted and ground truth distributions.

The training process spans 100 epochs, employing the AdamW optimizer. We employ a mean squared error (MSE) loss per patch to calculate the prediction error. The training halts if the loss on the validation set shows no improvement for at least 15 consecutive epochs. A warmup cosine schedule is utilized for the learning rate, with a warmup period of 20 epochs. The base learning rate (\(base\_lr\)) is set to \(1 \times 10^{-4}\), and the absolute learning rate (\(absolute\_lr\)) is calculated using the formula
\begin{equation*}
    absolute\_lr = base\_lr \times \frac{total\_batch\_size}{256}\,.
\end{equation*} 
A weight decay of $0.3$ is applied to the optimizer. Moreover, the
training is conducted on two \emph{NVIDIA RTX A5000} GPUs using
PyTorch's \emph{"Distributed Data Parallel"} to leverage distributed
training.  This configuration enhances the scalability and speed of
our training process.

For cross-validation, we employ a leave-one-out strategy with semantic maps. Our dataset is divided into training and validation maps using an 80/20 split. During each iteration, we exclude one map and train the model on the training set, validating on the validation set and evaluating on the withheld map for assessment. This process is repeated for each map in the dataset.
This strategy helps assess the generalization capability of our model across different semantic maps.
Additionally, we experiment with various patches and crop sizes to identify optimal configurations.
\section{Ablation Study}
\label{sec:ablation}

In order to methodically examine the effects of various elements in
the suggested Semantic Map-Aware Pedestrian Prediction 2 ({\tt
  semapp2}) model, we carry out an ablation study in this section. Our
objective is to comprehend the role that each component plays in the
overall performance and to choose the best performing model structure.
To measure the impact of each ablation, we employ quantitative metrics
such as KL divergence, reverse KL divergence and Earth Mover's
Distance. These distances provide insights into the model's ability to
accurately predict priors in semantic maps.

\subsection{Architectural Components}

To gauge the significance of specific architectural components, we
conducted a series of experiments, systematically tweaking key
elements within our {\tt semapp2} model based on Vision Transformer
(ViT).  First of all we need to choose the backbone for the
architecture between ViT-Base, ViT-Large and ViT-Huge~\cite{vit}. Then
we explore variations on patch dimensions and crop size with the
overarching goal of pinpointing the optimal configuration that strikes
a balance between model complexity and predictive accuracy.

\subsubsection{Backbone}

We start by investigating the impact of different backbones on the
{\tt MAE-semapp2} model, whose results are detailed in
Table~\ref{tab:backbone}.  During the ablation tests, we keep
unchanged the mask ratio of $75\%$, crop size of $64\times 64$ pixels
and patch size of $8\times 8$ pixels.
%-%
\begin{table}[t]
\footnotesize
  \centering
  \caption{Impact of the Backbone on {\tt semapp2} Model}
  \label{tab:backbone}
  \begin{tabular}{cccc}
    \toprule
    Backbone & KL-div & rKL-div & EMD \\
    \midrule
    ViT-Base  & $0.42\pm0.13$  & $2.52\pm1.88$  & $54.53\pm30.80$ \\
    ViT-Large  & $0.34\pm0.21$ & $2.19\pm1.84$ & $45.77\pm30.74$ \\
    \rowcolor{green!20}
    \textbf{ViT-Huge}  & \bm{$0.31\pm0.15$}  & \bm{$1.69\pm1.11$}  & \bm{$39.64\pm30.16$}  \\
    \bottomrule
  \end{tabular}
\end{table}
%-%
Our investigation revealed that the ViT-Huge backbone achieved the
best results, demonstrating lower values across KL-divergence, reverse
KL-divergence, and EMD metrics. Despite this superior performance, we
opt for utilizing the ViT-Large backbone for practical
considerations. The increment in performance with ViT-Huge is not
significant, and it does not justify the significantly longer training
times associated with its use. Thus, ViT-Large, being both proficient
and quicker to train, emerges as the pragmatic choice for our {\tt
  MAE-semapp2} model.  Moreover, as stated in~\cite{mae}, a
single-block decoder can perform strongly and speed up training, for
this reason we use a modified version of the ViT-Large changing the
decoder layers depth to 1.

\subsubsection{Crop Size}

Delving into the impact of varying the size of the analyzed crop of
the semantic map in our {\tt semapp2} model, we systematically
adjusted the crop size, resulting the $64$ the most promising
(see~\ref{tab:crop_size}).
%-%
\begin{table}[t]
\footnotesize
  \centering
  \caption{Impact of the Crop Size on {\tt semapp2} Model}
  \label{tab:crop_size}
  \begin{tabular}{cccc}
    \toprule
    Crop Size & KL-div & rKL-div & EMD \\
    \midrule
    32  & $0.62\pm0.18$  & $3.74\pm1.13$  & $54.56\pm29.84$  \\
    \rowcolor{green!20}
    \textbf{64}  & \bm{$0.34\pm0.21$} & \bm{$2.19\pm1.84$} & \bm{$45.77\pm30.74$}  \\
    100 & $0.56\pm0.19$  & $3.60\pm1.77$  & $117.38\pm93.33$  \\
    \bottomrule
  \end{tabular}
\end{table}
%-%

\subsubsection{Patch size}

Examining the influence of patch size on the {\tt semapp2} model, we
conducted experiments to observe variations in
performance. Table~\ref{tab:patch_size} shows how a patch size of 8
fits the paper needs.
%-%
\begin{table}[t]
\footnotesize
  \centering
  \caption{Impact of the Patch Size on {\tt semapp2} Model}
  \label{tab:patch_size}
  \begin{tabular}{cccc}
    \toprule
    Patch Size & KL-div & rKL-div & EMD \\
    \midrule
    % 4  & $0.50\pm0.17$  & $2.42\pm0.79$  & $55.11\pm31.38$  \\
    \rowcolor{green!20}
    \textbf{8}  & \bm{$0.34\pm0.21$} & \bm{$2.19\pm1.84$} & \bm{$45.77\pm30.74$}  \\
    16 & $0.52\pm0.10$  & $2.43\pm1.02$  & $53.02\pm34.39$  \\
    32 & $0.60\pm0.20$ & $4.57\pm1.79$ & $46.69\pm30.03$  \\
    \bottomrule
  \end{tabular}
\end{table}
%-%

\subsubsection{MAE Masking Percentage}

To investigate the impact of different masking percentages on the
performance of the {\tt semapp2} model, we conducted ablation
experiments by varying the masking percentage during training
obtaining the results of Table~\ref{tab:masking_ablation}. 
%-%
\begin{table}[t]
\footnotesize
    \centering
    \caption{Impact of Masking Percentage on {\tt semapp2} Model}
    \label{tab:masking_ablation}
    \begin{tabular}{cccc}
        \toprule
        Masking Ratio & KL-div & rKL-div & EMD \\
        \midrule
        0\% & $0.46\pm0.16$ & $2.19\pm1.50$ & $27.65\pm19.89$ \\
        25\% & $0.45\pm0.17$ & $2.32\pm1.66$ & $38.78\pm31.72$ \\
        50\% & $0.41\pm0.11$ & $2.36\pm1.44$ & $49.30\pm29.92$ \\
        \rowcolor{green!20}
        \textbf{75\%} & \bm{$0.34\pm0.21$} & \bm{$2.19\pm1.84$} & \bm{$45.77\pm30.74$} \\
        \bottomrule
    \end{tabular}
\end{table}
%-%
The masking percentage determines the proportion of patches excluded
during the training process, influencing the model's ability to
capture underlying patterns in the data.

% \subsection{Training Strategies}

% In addition to architectural components, we investigate different training strategies and techniques employed during the training of semapp2.

% \subsubsection{Loss Function Sensitivity}

% Our exploration dived into the sensitivity of semapp to different loss functions, particularly those tied to occupancy priors prediction. This analysis aimed to identify a loss function aligning with the specific objectives of our occupancy prediction task.

% \subsubsection{Learning Rate Sensitivity}

% The learning rate is a key hyperparameter affecting the convergence and generalization of the model. We conduct experiments with varying learning rates to understand their impact on training stability and predictive performance.

% \subsubsection{Data Augmentation Effects}

% To assess semapp's robustness, we investigated the influence of diverse data augmentation techniques. Altering augmentation strategies during training, we scrutinized their effects on model generalization and performance across varied scenarios.
\section{Results and Discussion}
\label{sec:results}
% Report the mean and standard deviations of KL-divergences for both ViT and semapp on the Stanford Drone Dataset. Highlight the performance differences, including qualitative improvements observed in semapp. Visualize the generalization capabilities of ViT and discuss the relation between the number of samples, performance, and inference time.

The evaluation metric involves computing the Kullback-Leibler (KL)
divergence, Reverse KL divergence and Earth Mover's Distance (EMD) for
all the leave-one-out maps, resulting in a mean metric value along
with standard deviation, providing insights into the model's
generalization across various semantic maps.

\begin{figure}[t] 
  \begin{minipage}[b]{0.5\linewidth}
    \raggedleft
    \includegraphics[width=0.62\linewidth]{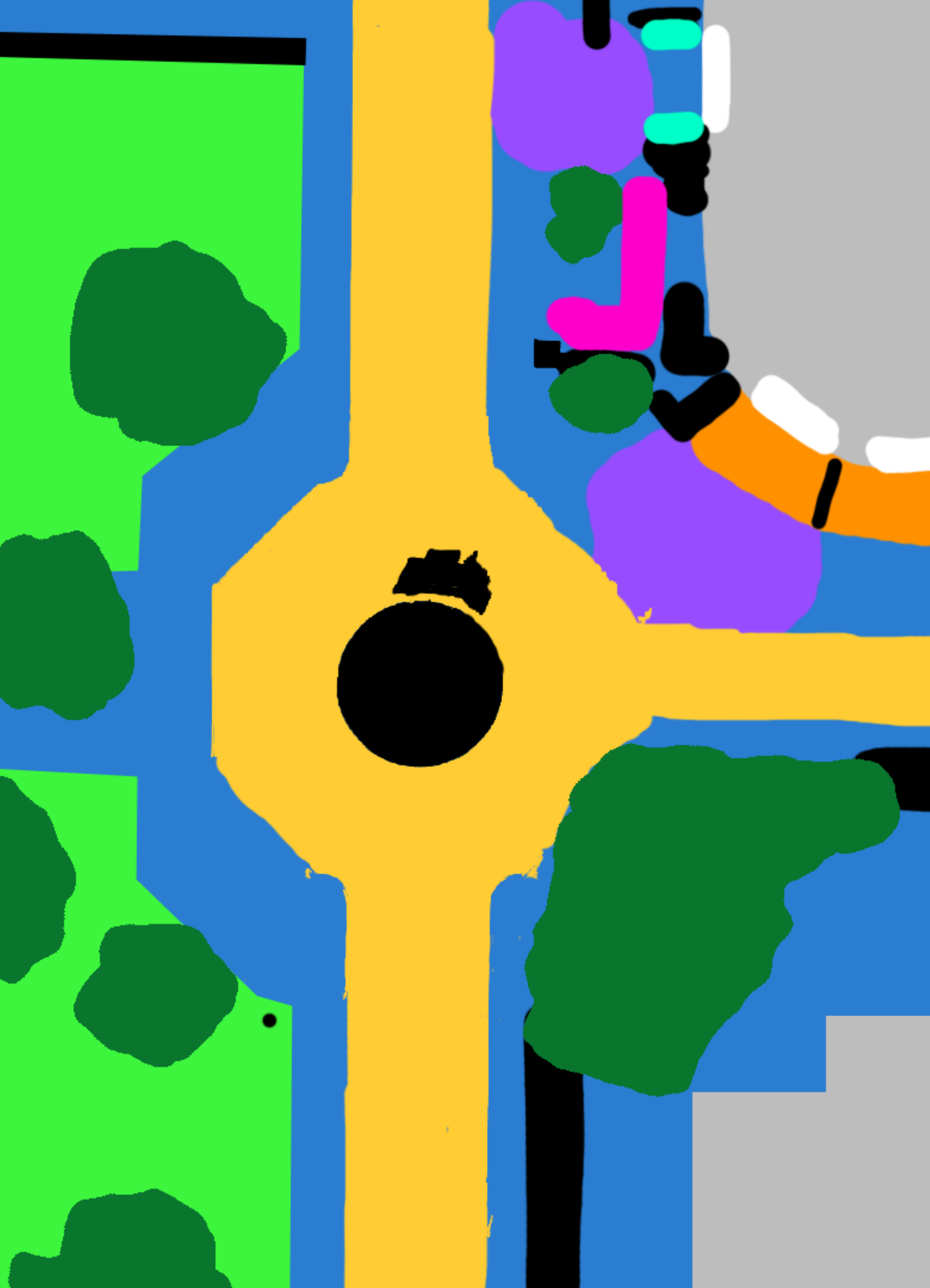} 
    \vspace{0.9ex}
  \end{minipage}%%
  \begin{minipage}[b]{0.5\linewidth}
    \raggedright
    \includegraphics[width=0.62\linewidth]{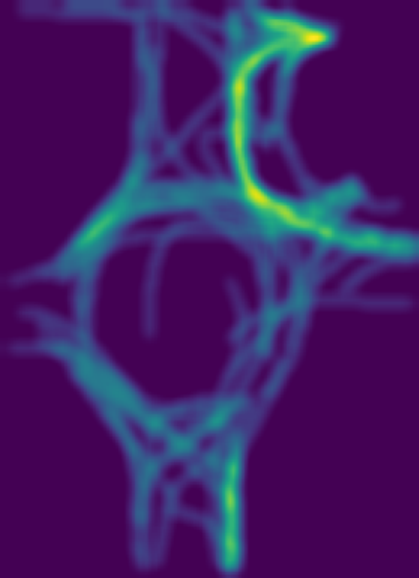} 
    \vspace{0.9ex}
  \end{minipage}
  \begin{minipage}[b]{0.33\linewidth}
    \centering
    \includegraphics[width=0.95\linewidth]{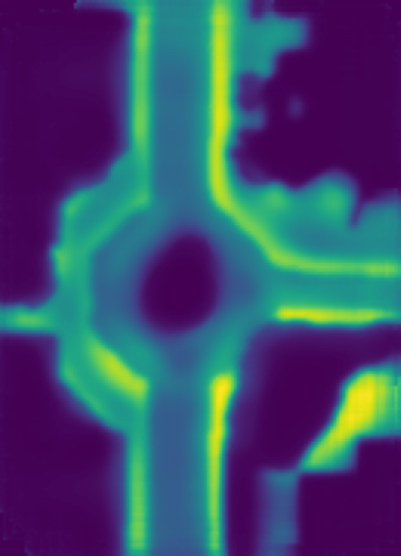} 
    \hspace{0.1ex}
  \end{minipage}%% 
  \begin{minipage}[b]{0.33\linewidth}
    \centering
    \includegraphics[width=0.95\linewidth]{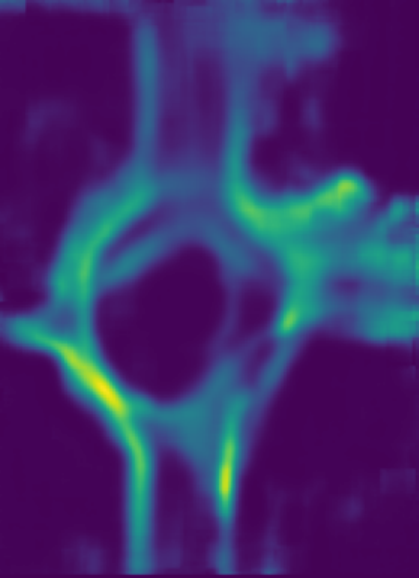} 
  \end{minipage} 
    \begin{minipage}[b]{0.33\linewidth}
    \centering
    \includegraphics[width=0.95\linewidth]{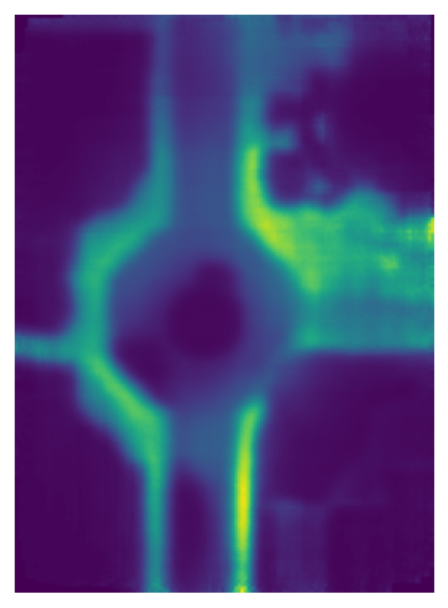}
    \label{fig:gates1_MAE}
  \end{minipage} 
  \caption{Qualitative comparison of results in the Stanford Drone
    Dataset. Our ViT-based model showcases competitive performance
    compared to {\tt semapp} (Rudenko et al. \cite{rudenko}),
    demonstrating the effectiveness of Vision Transformers in
    predicting occupancy priors. \textbf{Top left}: presents the
    original semantic map highlighting different classes, \textbf{Top
      right}: displays the ground-truth distribution of
    occupancies. \textbf{Bottom left}, \textbf{Bottom middle} and
    \textbf{Bottom right} showcase the predictions generated by {\tt
      semapp}, {\tt semapp2} and {\tt MAE-semapp2}, respectively.}
  \label{fig:semantic_comparison}
\end{figure}

\begin{figure}[t] 
  \begin{minipage}[b]{0.5\linewidth}
    \raggedleft
    \includegraphics[width=0.7\linewidth]{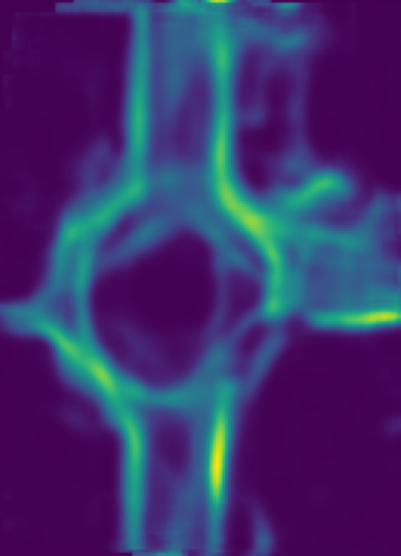} 
    \vspace{0.8ex}
  \end{minipage}%%
  % \hspace{2ex}
  \begin{minipage}[b]{0.5\linewidth}
    \raggedright
    \includegraphics[width=0.7\linewidth]{images/gates1_semapp2_13labels} 
    \vspace{0.8ex}
  \end{minipage}
  \caption{Qualitative comparison between using 9 semantic classes (\textbf{Left}) and 13 semantic classes (\textbf{Right})}
  \label{fig:comp_labelsN}
\end{figure}

We present a qualitative comparative analysis in
Figure~\ref{fig:semantic_comparison} between {\tt semapp} and {\tt
  semapp2}.
%-%
%-%
The top-left section depicts the semantic map and the top-right
section represents the corresponding ground-truth occupancy
distribution. In the bottom-left, predictions from {\tt semapp}, while
the bottom-centre and the bottom-right show predictions from {\tt
  semapp2} and {\tt MAE-semapp2}. Moreover, in
Figure~\ref{fig:comp_labelsN}, we compare the quality of the
predictions of the model {\tt semapp2} using 9 semantic labels, shown
on the left in the figure, or using 13 semantic labels, on the right.
%-%
%-%
Visually the difference is barely noticeable, but quantitatively we have
a slight improvement, as reported in Tables~\ref{tab:semantic_influence_9} and \ref{tab:semantic_influence_13}.

\subsection{Quantitative Evaluation}

We provide in Table~\ref{tab:quantitative_comp} the mean and standard
deviations of KL-divergences, reverse KL-divergences and EMDs for all
three models ({\tt semapp}, {\tt semapp2}, {\tt MAE-semapp2}) applied
to the Stanford Drone Dataset using a cross-validation approach, as
described in Section~\ref{sec:datasets}.
%-%
\begin{table}[t]
\footnotesize
  \centering
  \caption{Stanford Drone Dataset Quantitative Evaluation}
  \label{tab:quantitative_comp}
  \begin{tabular}{lccc}
    \toprule
    Method & Average KL-Div & Average rKL-Div & Average EMD \\
    \midrule
    {\tt semapp} & $0.58\pm0.14$ & $2.43\pm1.24$ & $41.16\pm26.98$ \\
    {\tt semapp2} & $0.46\pm0.16$ & $2.19\pm1.50$ & $27.65\pm19.89$ \\
    {\tt MAE-semapp2} & $0.34\pm0.21$ & $2.19\pm1.84$ & $45.77\pm30.74$ \\
    \bottomrule
  \end{tabular}
\end{table}
%-%
In the Stanford Drone Dataset, {\tt semapp2} shows competitive
performance compared to {\tt semapp}. This main comparison provides
insights into the effectiveness of ViT in predicting occupancy priors
based on semantic information.

\subsubsection*{semapp2 vs. MAE-semapp2}

\begin{figure}[t]
    \centering
    \includegraphics[width=\linewidth]{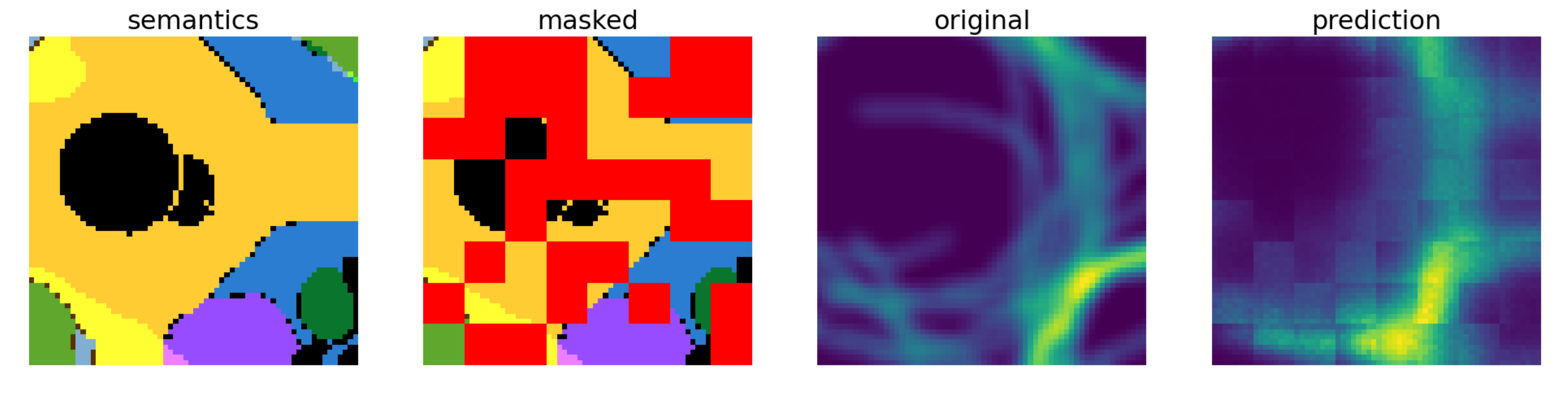}
    \caption{Example of prediction using the MAE-based semapp2.}
    \label{fig:MAEex}
\end{figure}

The four images in Fig~\ref{fig:MAEex} showcase different aspects of
the prediction process using the MAE-based {\tt semapp2}.
%-%
%-%
The first {\em semantics} image represents the original crop of the semantic map,
while the second {\em masked} image displays the same crop
after the masking process, emphasizing the regions of interest during
the model's inference. The {\em original} image presents the ground
truth of the occupancy distribution, providing a reference for the
expected outcome. Finally, the {\em prediction} image depicts the
result of the MAE-based {\tt semapp2}, illustrating the model's
capability to anticipate and reproduce the occupancy distribution
based on the masked semantic input.

\begin{figure*}[t] 
  \begin{minipage}[b]{0.25\linewidth}
    \centering
    \includegraphics[width=0.95\linewidth]{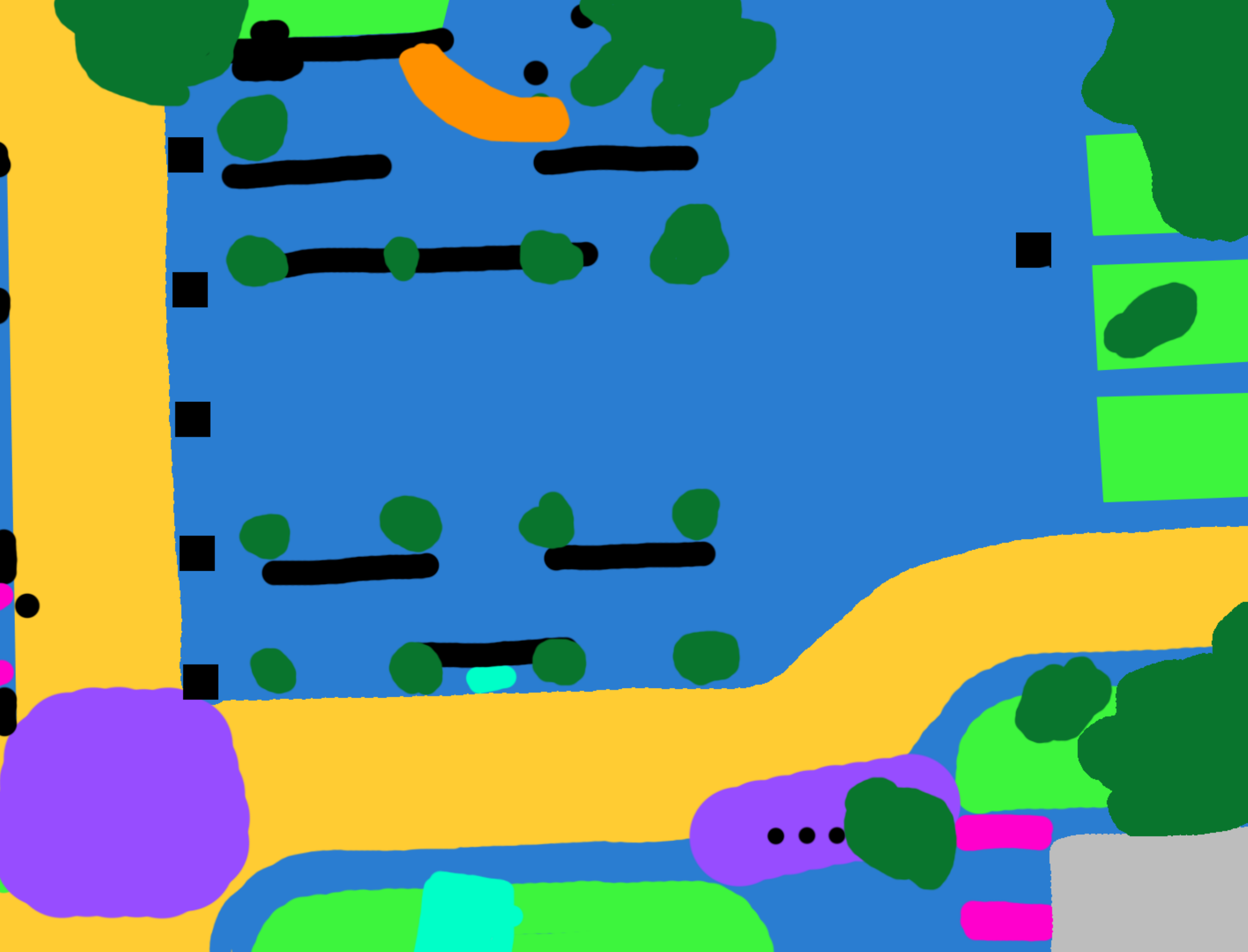} 
    \vspace{1.5ex}
  \end{minipage}%%
  \begin{minipage}[b]{0.25\linewidth}
    \centering
    \includegraphics[width=0.95\linewidth]{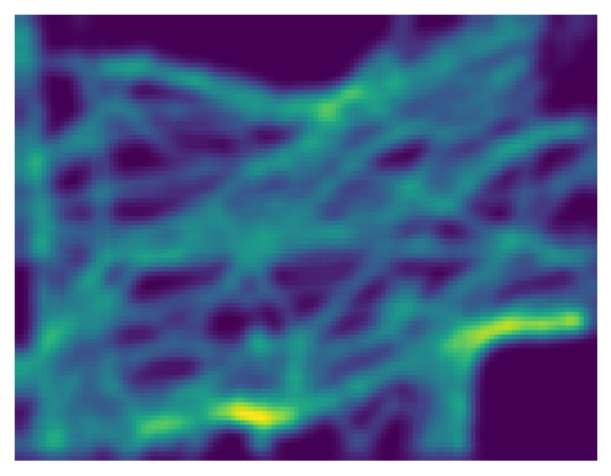} 
    \vspace{1.5ex}
  \end{minipage}%%
  \begin{minipage}[b]{0.25\linewidth}
    \centering
    \includegraphics[width=0.95\linewidth]{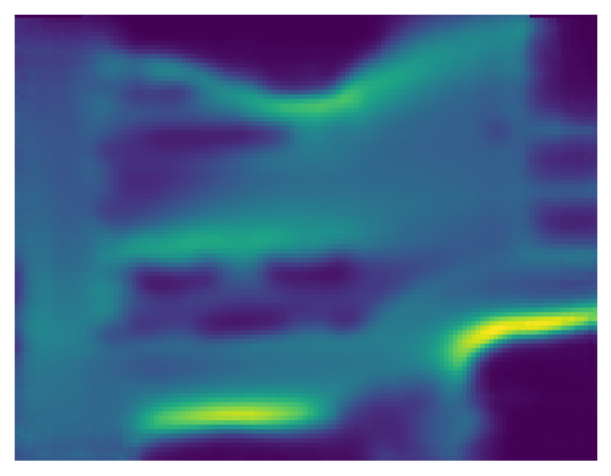} 
    \vspace{1.5ex}
  \end{minipage}%%
  \begin{minipage}[b]{0.25\linewidth}
    \centering
    \includegraphics[width=0.95\linewidth]{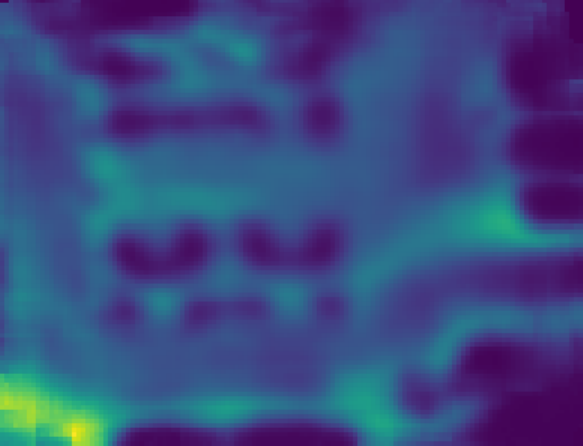} 
    \vspace{1.5ex}
  \end{minipage}
  \begin{minipage}[b]{0.25\linewidth}
    \centering
    \includegraphics[width=0.95\linewidth]{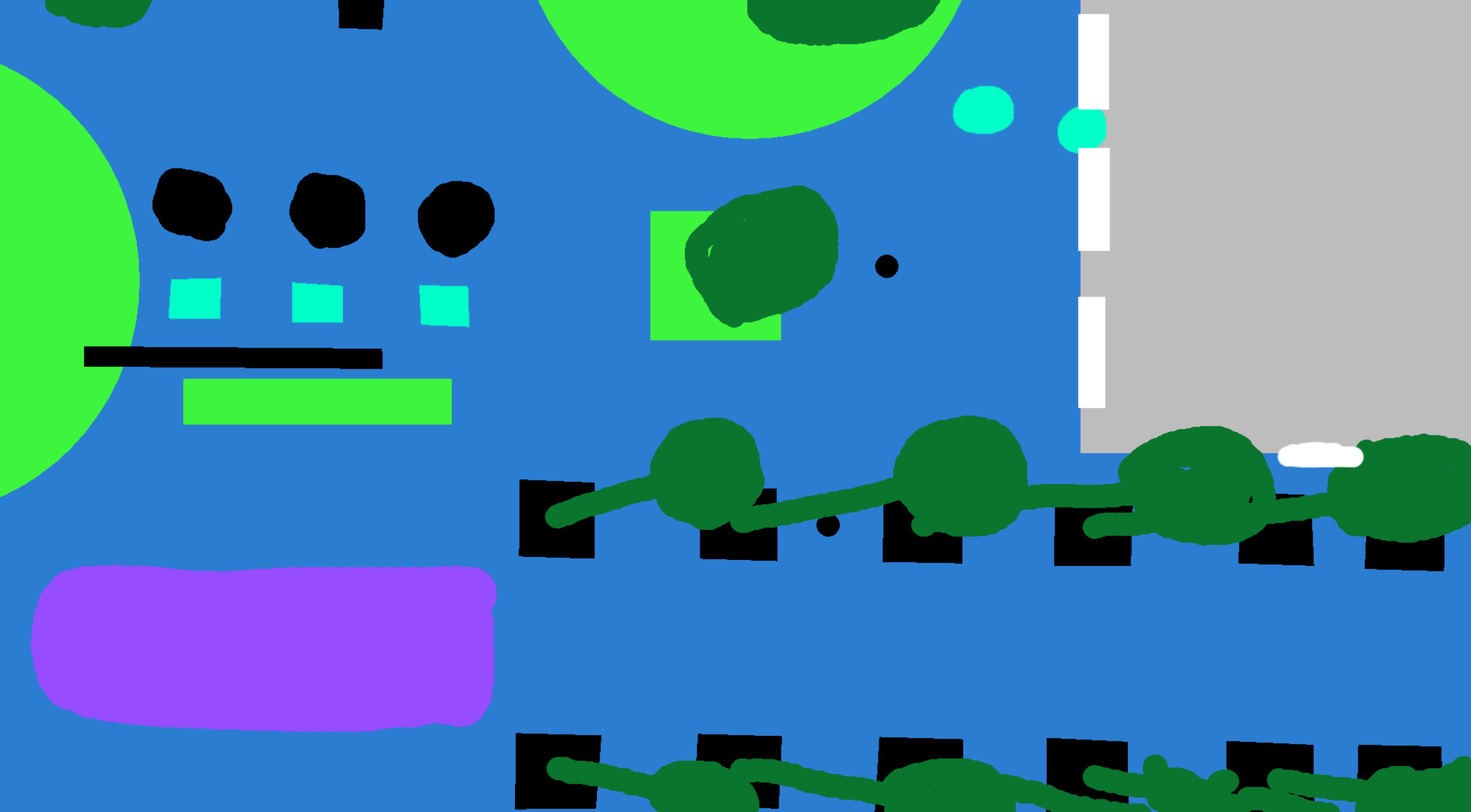} 
  \end{minipage}%%
  \begin{minipage}[b]{0.25\linewidth}
    \centering
    \includegraphics[width=0.95\linewidth]{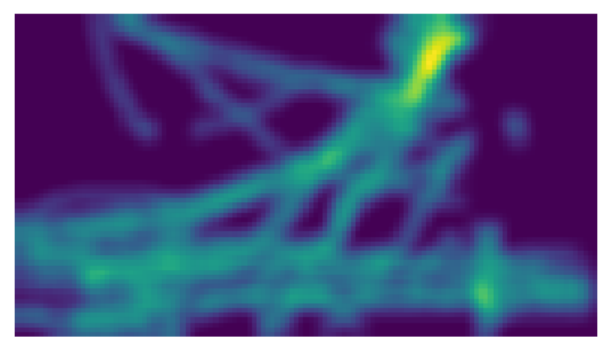} 
  \end{minipage}%%
  \begin{minipage}[b]{0.25\linewidth}
    \centering
    \includegraphics[width=0.95\linewidth]{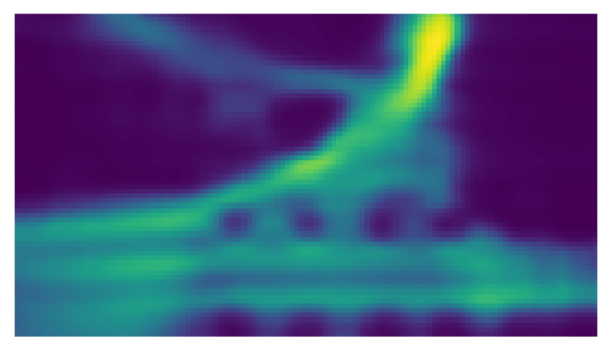}
  \end{minipage}%%
  \begin{minipage}[b]{0.25\linewidth}
    \centering
    \includegraphics[width=0.95\linewidth]{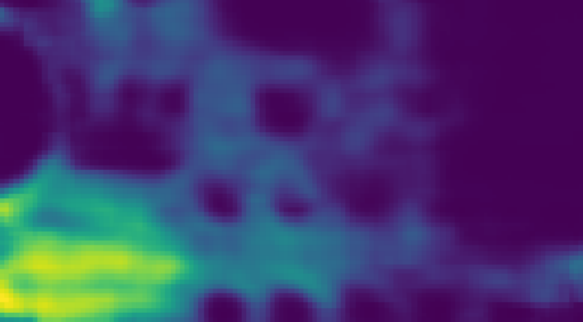}
  \end{minipage} 
  \caption{Qualitative evaluation of the predictions of the {\tt
      MAE-semapp2}. The \textbf{First column} shows the semantic maps,
    the \textbf{Second column} shows the groundtruths while the
    \textbf{Third column} shows the predictions using the MAE-based
    {\tt semapp2}. The \textbf{Fourth column} shows the predictions
    using the ViT-based {\tt semapp2}. (First row depicts the map
    bookstore, video 0, and the second row the map coupa, video 3)}
  \label{fig:MAE_comp}
\end{figure*}

The MAE-based {\tt semapp2} exhibits a notable level of generalization
ability compared to the ViT-based {\tt semapp2}. In~\cite{mae}, the
authors demonstrate that masking patches in MAEs does not result in a
decremental impact on reconstruction and classification, underscoring
the significant data redundancy present in vision tasks. This
observation suggests that the MAE model might be well-suited for
learning underlying laws of social motion.  Indeed, in
Table~\ref{tab:quantitative_comp}, the {\tt MAE-semapp2} variation
shows slightly worse performance over the EMD metric compared to both
{\tt semapp} and {\tt semapp2}. However, from a qualitative
evaluation, the model seems to predict extremely well local variations
of the occupancy distribution.  Two examples are shown in
Figure~\ref{fig:MAE_comp}: the MAE-based {\tt semapp2} is clearly
superior at predicting the distribution.
%-%
%-%
The low metric values could be due to a lack of trajectories in the
timespan analysed in the SDD video. For this reason, the
generalization ability of the MAE, especially in complex scenarios,
warrants further exploration and investigation in future works.  To
assess the quality of the prediction, it could be necessary to collect
more data on a specific location at different times in order to
converge to a global probability distribution of the human occupancy,
rather than the time-variant distribution that we obtain from the SDD
videos.

\subsection{Predicting Stops and Velocities}

\begin{figure}[t]
  \begin{minipage}[b]{\linewidth}
    \centering
    \includegraphics[width=0.5\linewidth]{images/stanford_coupa3_colors} 
    \vspace{0.8ex}
  \end{minipage}
  \begin{minipage}[b]{0.5\linewidth}
    \raggedleft
    \includegraphics[width=0.9\linewidth]{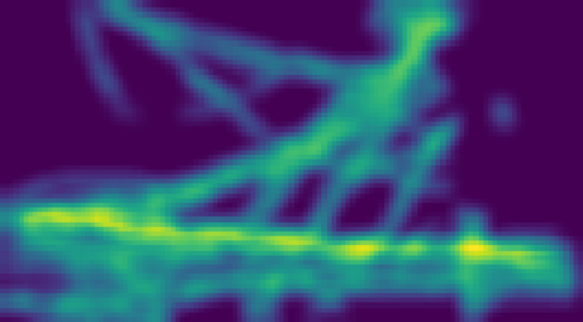} 
    \vspace{0.8ex}
  \end{minipage}%%
  % \hspace{2ex}
  \begin{minipage}[b]{0.5\linewidth}
    \raggedright
    \includegraphics[width=0.9\linewidth]{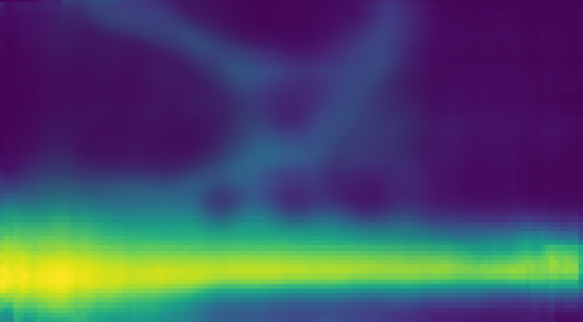}
    \vspace{0.8ex}
  \end{minipage}
  \begin{minipage}[b]{0.5\linewidth}
    \raggedleft
    \includegraphics[width=0.9\linewidth]{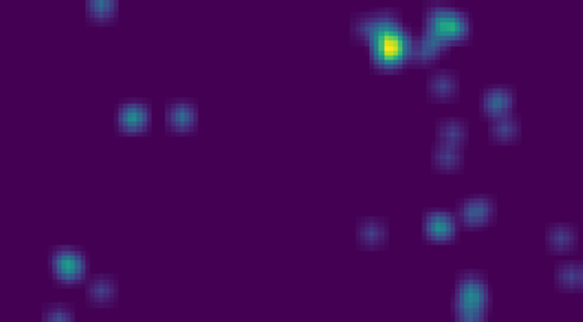} 
    \vspace{0.8ex}
  \end{minipage}%%
  % \hspace{2ex}
  \begin{minipage}[b]{0.5\linewidth}
    \raggedright
    \includegraphics[width=0.9\linewidth]{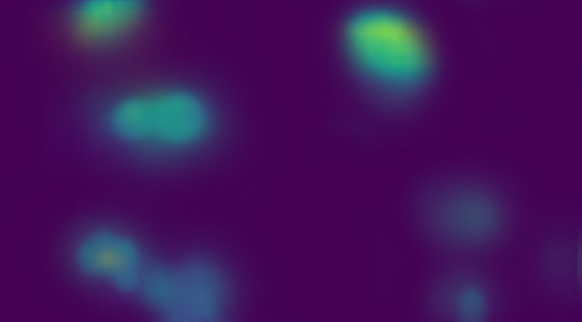}
    \vspace{0.8ex}
  \end{minipage}
  \caption{Prediction of the velocity profile distribution and of the stops distribution in the coupa map.}
  \label{fig:velstops}
\end{figure}

Furthermore, we delve into assessing the network's proficiency in
predicting stops and velocities—priors that have been relatively
underexplored in existing literature. This unique investigation holds
significant implications for the field of mobility, where accurately
anticipating stops and velocities could be crucial for enhancing
planning tasks. Our exploration of these nuanced prediction tasks adds
valuable insights to the broader understanding of Vision Transformers'
capabilities in addressing complex aspects of occupancy prediction,
particularly in real-world mobility scenarios.
Figures~\ref{fig:velstops} present the prediction of the velocity
profile distribution and the stop distribution.
%-%
%-%
Additionally, Table~\ref{tab:velstops} provides a quantitative
evaluation of the predictions of velocities and stops.
%-%
\begin{table}[t]
\footnotesize
\centering
\caption{Quantitative Evaluation of Velocities and Stops}
\label{tab:velstops}
\begin{tabular}{lccc}
\toprule
& KL-div & rKL-div & EMD \\
\midrule
Velocities & $0.47\pm0.15$ & $2.50\pm1.51$ & $40.18\pm26.55$ \\
Stops & $0.63\pm0.15$ & $2.15\pm1.20$ & $52.88\pm27.94$ \\
\bottomrule
\end{tabular}
\end{table}
%-%

% \pla{Show how added semantic classes improve the stop map.}

% \pla{Insert experiment of closed, semiclosed and open passage in corridor of map}
\section{Conclusion}
\label{sec:conclusions}
The analysis of human occupancy of the different areas of an
environment is essential to enable safe and efficient navigation of
mobile robots. The past literature has shown that the use of semantic
maps can significantly speed up the reconstruction of this information
by breaking down the environment into several smaller parcels. The
price to pay is a potential limitation of the accuracy due to the use
of local information. We have proposed a solution that holds the
promise to mitigate this problem by the use of a ViT backbone. Indeed
the use of transformers allows the network to learn the spatial
relation between nearby areas, hence reconstructing a global view of
the environment.  The results show that our solution significantly
improves the prediction accuracy with a limited impact on the
computation time, which remains acceptable for real--time applications
of the solution.  Many problems remain open and are reserved for
future research activities.  The first activity will be to test our
method on a more complete dataset than the Stanford Drone dataset used
for this paper, which contains more trajectories per map, to better
evaluate the generalisation ability of the models.  A second activity
will be the integration of the module into a robot navigation
framework. Third, we are considering a possible extension of the
approach to predicting the motion of bicycles or cars, which would
open other interesting application opportunities (i.e., autonomous
driving). Finally, we are actually working to extend the same idea to
cobots in production cells for product quality control and reworking
of defected working pieces.

\section{Acknowledgments}
Co-funded by the European Union. Views and opinions expressed are
however those of the author(s) only and do not necessarily reflect
those of the European Union or the European Commission. Neither the
European Union nor the granting authority can be held responsible for
them (EU - HE Magician – Grant Agreement 101120731). Moreover, this
work was partially funded by the European Commission grant number
101016906 (Project CANOPIES).

%%%%%%%%%%%%%%%%%%%%%%%
%%%%%%%%%%%%%%%%%%%%%%%
%%%%%%%%%%%%%%%%%%%%%%%

\bibliographystyle{IEEEtran}
\bibliography{biblio}

\begin{thebibliography}{10}
\providecommand{\url}[1]{#1}
\csname url@samestyle\endcsname
\providecommand{\newblock}{\relax}
\providecommand{\bibinfo}[2]{#2}
\providecommand{\BIBentrySTDinterwordspacing}{\spaceskip=0pt\relax}
\providecommand{\BIBentryALTinterwordstretchfactor}{4}
\providecommand{\BIBentryALTinterwordspacing}{\spaceskip=\fontdimen2\font plus
\BIBentryALTinterwordstretchfactor\fontdimen3\font minus
  \fontdimen4\font\relax}
\providecommand{\BIBforeignlanguage}[2]{{%
\expandafter\ifx\csname l@#1\endcsname\relax
\typeout{** WARNING: IEEEtran.bst: No hyphenation pattern has been}%
\typeout{** loaded for the language `#1'. Using the pattern for}%
\typeout{** the default language instead.}%
\else
\language=\csname l@#1\endcsname
\fi
#2}}
\providecommand{\BIBdecl}{\relax}
\BIBdecl

\bibitem{Singamaneni2024}
P.~T. Singamaneni, P.~Bachiller-Burgos, L.~J. Manso, A.~Garrell, A.~Sanfeliu,
  A.~Spalanzani, and R.~Alami, ``Advances and challenges in human-aware social
  robot navigation: A survey,'' \emph{International Journal Robotics Research},
  February 2024.

\bibitem{rudenko}
A.~Rudenko, L.~Palmieri, J.~Doellinger, A.~J. Lilienthal, and K.~O. Arras,
  ``Learning occupancy priors of human motion from semantic maps of urban
  environments,'' \emph{IEEE Robotics and Automation Letters}, vol.~6, no.~2,
  pp. 3248--3255, 2021.

\bibitem{mavrogiannis2023core}
C.~Mavrogiannis, F.~Baldini, A.~Wang, D.~Zhao, P.~Trautman, A.~Steinfeld, and
  J.~Oh, ``Core challenges of social robot navigation: A survey,'' \emph{ACM
  Transactions on Human-Robot Interaction}, vol.~12, no.~3, pp. 1--39, 2023.

\bibitem{rudenko2020human}
A.~Rudenko, L.~Palmieri, M.~Herman, K.~M. Kitani, D.~M. Gavrila, and K.~O.
  Arras, ``Human motion trajectory prediction: a survey,'' \emph{The
  International Journal of Robotics Research}, vol.~39, no.~8, pp. 895--935,
  2020.

\bibitem{FarinaFGGP17}
F.~Farina, D.~Fontanelli, A.~Garulli, A.~Giannitrapani, and D.~Prattichizzo,
  ``{Walking Ahead: The Headed Social Force Model},'' \emph{PLOS ONE}, vol.~12,
  no.~1, pp. 1--23, 1 2017.

\bibitem{7535484}
S.~Zernetsch, S.~Kohnen, M.~Goldhammer, K.~Doll, and B.~Sick, ``Trajectory
  prediction of cyclists using a physical model and an artificial neural
  network,'' in \emph{2016 IEEE Intelligent Vehicles Symposium (IV)}, 2016, pp.
  833--838.

\bibitem{7995734}
E.~A.~I. Pool, J.~F.~P. Kooij, and D.~M. Gavrila, ``Using road topology to
  improve cyclist path prediction,'' in \emph{2017 IEEE Intelligent Vehicles
  Symposium (IV)}, 2017, pp. 289--296.

\bibitem{vemula2017modeling}
A.~Vemula, K.~Muelling, and J.~Oh, ``Modeling cooperative navigation in dense
  human crowds,'' 2017.

\bibitem{AntonucciPBPF21access}
A.~Antonucci, G.~R. Papini, P.~Bevilacqua, L.~Palopoli, and D.~Fontanelli,
  ``{Efficient Prediction of Human Motion for Real-Time Robotics Applications
  with Physics-inspired Neural Networks},'' \emph{IEEE Access}, vol.~10, pp.
  144--157, December 2021.

\bibitem{kaleci2020semantic}
B.~Kaleci, {\c{C}}.~M. {\c{S}}enler, H.~Duta{\u{g}}ac{\i}, and O.~Parlaktuna,
  ``Semantic classification of mobile robot locations through 2d laser scans,''
  \emph{Intelligent Service Robotics}, vol.~13, no.~1, pp. 63--85, 2020.

\bibitem{cnn}
A.~Krizhevsky, I.~Sutskever, and G.~E. Hinton, ``Imagenet classification with
  deep convolutional neural networks,'' in \emph{Advances in neural information
  processing systems}, 2012, pp. 1097--1105.

\bibitem{minaee2021image}
S.~Minaee, Y.~Boykov, F.~Porikli, A.~Plaza, N.~Kehtarnavaz, and D.~Terzopoulos,
  ``Image segmentation using deep learning: A survey,'' \emph{IEEE transactions
  on pattern analysis and machine intelligence}, vol.~44, no.~7, pp.
  3523--3542, 2021.

\bibitem{doellinger}
J.~Doellinger, M.~Spies, and W.~Burgard, ``Predicting occupancy distributions
  of walking humans with convolutional neural networks,'' \emph{IEEE Robotics
  and Automation Letters}, vol.~3, no.~3, pp. 1522--1528, 2018.

\bibitem{densecnn}
S.~Jégou, M.~Drozdzal, D.~Vazquez, A.~Romero, and Y.~Bengio, ``The one hundred
  layers tiramisu: Fully convolutional densenets for semantic segmentation,''
  2017.

\bibitem{denseblock}
G.~Huang, Z.~Liu, L.~van~der Maaten, and K.~Q. Weinberger, ``Densely connected
  convolutional networks,'' 2018.

\bibitem{srivastava2023omnivec}
S.~Srivastava and G.~Sharma, ``Omnivec: Learning robust representations with
  cross modal sharing,'' 2023.

\bibitem{wortsman2022model}
M.~Wortsman, G.~Ilharco, S.~Y. Gadre, R.~Roelofs, R.~Gontijo-Lopes, A.~S.
  Morcos, H.~Namkoong, A.~Farhadi, Y.~Carmon, S.~Kornblith, and L.~Schmidt,
  ``Model soups: averaging weights of multiple fine-tuned models improves
  accuracy without increasing inference time,'' 2022.

\bibitem{vit}
A.~Dosovitskiy, L.~Beyer, A.~Kolesnikov, D.~Weissenborn, X.~Zhai,
  T.~Unterthiner, M.~Dehghani, M.~Minderer, G.~Heigold, S.~Gelly, J.~Uszkoreit,
  and N.~Houlsby, ``An image is worth 16x16 words: Transformers for image
  recognition at scale,'' 2021.

\bibitem{strudel2021segmenter}
R.~Strudel, R.~Garcia, I.~Laptev, and C.~Schmid, ``Segmenter: Transformer for
  semantic segmentation,'' 2021.

\bibitem{vaswani2023attention}
A.~Vaswani, N.~Shazeer, N.~Parmar, J.~Uszkoreit, L.~Jones, A.~N. Gomez,
  L.~Kaiser, and I.~Polosukhin, ``Attention is all you need,'' 2023.

\bibitem{devlin2019bert}
J.~Devlin, M.-W. Chang, K.~Lee, and K.~Toutanova, ``Bert: Pre-training of deep
  bidirectional transformers for language understanding,'' 2019.

\bibitem{brown2020language}
T.~B. Brown, B.~Mann, N.~Ryder, M.~Subbiah, J.~Kaplan, P.~Dhariwal,
  A.~Neelakantan, P.~Shyam, G.~Sastry, A.~Askell, S.~Agarwal, A.~Herbert-Voss,
  G.~Krueger, T.~Henighan, R.~Child, A.~Ramesh, D.~M. Ziegler, J.~Wu,
  C.~Winter, C.~Hesse, M.~Chen, E.~Sigler, M.~Litwin, S.~Gray, B.~Chess,
  J.~Clark, C.~Berner, S.~McCandlish, A.~Radford, I.~Sutskever, and D.~Amodei,
  ``Language models are few-shot learners,'' 2020.

\bibitem{mae}
K.~He, X.~Chen, S.~Xie, Y.~Li, P.~Dollár, and R.~Girshick, ``Masked
  autoencoders are scalable vision learners,'' 2021.

\bibitem{kl_divergence_paper}
S.~Kullback and R.~A. Leibler, ``On information and sufficiency,'' \emph{The
  Annals of Mathematical Statistics}, vol.~22, no.~1, pp. 79--86, 1951.

\bibitem{rubner2000earth}
Y.~Rubner, C.~Tomasi, and L.~J. Guibas, ``The earth mover's distance as a
  metric for image retrieval,'' \emph{International Journal of Computer
  Vision}, vol.~40, pp. 99--121, November 2000.

\bibitem{Robicquet2016}
A.~Robicquet, A.~Sadeghian, A.~Alahi, and S.~Savarese, ``Learning social
  etiquette: Human trajectory prediction in crowded scenes,'' in \emph{European
  Conference on Computer Vision (ECCV)}, 2016.

\end{thebibliography}

\end{document}